\documentclass[10pt]{article}
\usepackage{indentfirst}   
\usepackage{graphicx}
\usepackage{amsmath, amssymb}
\usepackage{booktabs}
\usepackage{multirow}
\usepackage{geometry}
\usepackage{float}
\usepackage{caption}
\usepackage{subcaption}
\usepackage{hyperref}
\title{A Lightweight Medical Image Classification Framework via Self-Supervised Contrastive Learning and Quantum-Enhanced Feature Modeling}

\author{
Jingsong Xia$^{1}$, Siqi Wang$^{1}$ \\
\small $^{1}$The Second Clinical Medical College, Nanjing Medical University \\
\small \texttt{xiajingsong2@gmail.com; wsq03925@163.com}
}

\date{January 2026}

\begin{document}
\maketitle

\begin{abstract}
\textbf{Background:} Intelligent analysis of medical images plays a critical role in clinical decision support systems. However, real-world medical imaging applications are often constrained by the scarcity of high-quality annotations, limited computational resources, and insufficient model generalization. Recently, self-supervised learning (SSL) and quantum machine learning (QML) have demonstrated promising potential in representation learning and feature enhancement, respectively. Nevertheless, their systematic integration and lightweight deployment in medical imaging remain underexplored.

\textbf{Methods:} We propose a lightweight medical image analysis framework that integrates self-supervised contrastive learning with quantum-enhanced classification. MobileNetV2 is adopted as the backbone network, and a SimCLR-style self-supervised contrastive learning paradigm is employed for unlabeled pretraining. On top of the classical feature extractor, a lightweight parameterized quantum circuit (PQC) is embedded as a quantum feature enhancement module, forming a hybrid classical--quantum representation learning architecture. The model is first pretrained on unlabeled medical images using contrastive objectives and subsequently fine-tuned on a limited number of labeled samples to improve downstream classification performance and generalization.

\textbf{Results:} Experimental results demonstrate that, with only approximately 2--3 million parameters and under low-computational-cost settings, the proposed framework achieves superior classification performance on the test set. The proposed method consistently outperforms baseline models without self-supervised learning or quantum enhancement in terms of Accuracy, AUC, and F1-score. Visualization analyses further confirm the improved discriminative capability and feature stability of the learned representations.

\textbf{Conclusion:} This study presents the first systematic integration of self-supervised contrastive learning and quantum-enhanced feature modeling in a lightweight medical image classification framework. The proposed approach offers a feasible and forward-looking solution for high-performance medical artificial intelligence under resource-constrained conditions.
\end{abstract}


\section{Introduction}

Medical imaging modalities, including X-ray, computed tomography (CT), magnetic resonance imaging (MRI), ultrasound, and coronary angiography, play an indispensable role in disease screening, diagnostic stratification, and treatment planning. With continuous advances in imaging hardware, modern clinical practice has generated a large volume of high-dimensional and highly complex medical image data, providing a solid foundation for artificial intelligence (AI)-assisted diagnosis. In recent years, deep learning methods, particularly convolutional neural networks (CNNs), have achieved remarkable success in medical image classification, segmentation, and detection tasks, demonstrating performance comparable to or even exceeding that of human experts in specific scenarios.

Despite these advances, the deployment of deep learning models in real-world clinical environments remains challenging~\cite{13,14}. First, high-quality medical image annotation relies heavily on experienced radiologists or clinical experts, leading to high labor costs, long annotation cycles, and unavoidable inter-observer variability. This issue is particularly pronounced in tasks involving angiography, functional imaging, and multimodal data, significantly limiting the effectiveness of large-scale supervised learning. Second, mainstream deep learning architectures, such as ResNet, DenseNet, and vision Transformer-based models, typically involve substantial parameter counts and computational complexity. These requirements hinder their deployment in resource-constrained settings, including primary healthcare institutions, portable medical devices, and research environments relying on standard laptops. Third, common characteristics of medical imaging datasets, such as class imbalance, small sample sizes, and cross-center distribution shifts, often result in degraded generalization performance, further restricting clinical applicability~\cite{15,16}.

In this context, self-supervised learning (SSL) has emerged as a promising strategy to alleviate annotation scarcity by leveraging unlabeled data to learn discriminative representations~\cite{1,2}. In particular, contrastive learning-based methods, including SimCLR~\cite{3}, MoCo~\cite{4}, and BYOL~\cite{5}, have achieved breakthrough performance in natural image analysis by maximizing representation consistency across multiple augmented views of the same instance. By formulating instance-level discrimination tasks, these methods enable models to learn well-structured embedding spaces without manual labels. However, medical images fundamentally differ from natural images in terms of imaging mechanisms, semantic structures, and information distributions. Directly applying contrastive augmentation strategies designed for natural images may disrupt critical anatomical structures or pathological patterns, thereby undermining self-supervised learning effectiveness. Consequently, designing contrastive learning strategies that respect medical imaging constraints and effectively transfer to downstream clinical tasks remains an open research challenge.

Meanwhile, quantum machine learning~\cite{6,8}, as an emerging interdisciplinary field at the intersection of quantum computing and artificial intelligence, has attracted increasing attention~\cite{7}. Theoretically, parameterized quantum circuits can exploit quantum superposition and entanglement to model complex nonlinear correlations in high-dimensional feature spaces, offering complementary representational capabilities to classical neural networks. Although current quantum hardware remains in its early stages, hybrid quantum--classical models implemented via classical simulators have demonstrated feasibility in various pattern recognition tasks~\cite{9,10}. Embedding lightweight quantum feature mapping modules into classical deep networks introduces a novel modeling perspective for medical image representation learning.

Motivated by these considerations, we propose a lightweight framework that integrates self-supervised contrastive learning with quantum-enhanced feature modeling for medical image classification~\cite{11,12}. Without imposing substantial computational overhead, the proposed method effectively leverages unlabeled medical images for representation pretraining and employs quantum-enhanced modules to capture higher-order nonlinear feature interactions. As a result, the framework achieves improved performance and generalization in small-sample and resource-constrained scenarios. The main contributions of this work are summarized as follows: (1) we design a self-supervised contrastive learning strategy tailored to the characteristics of medical images to mitigate annotation scarcity; (2) we develop a lightweight quantum feature enhancement module integrated via residual fusion to ensure training stability and computational feasibility; and (3) we validate the effectiveness and efficiency of the proposed framework through comprehensive medical image classification experiments.

\section{Methods}

\subsection{Overall Framework and Two-Stage Training Paradigm}

Medical image artificial intelligence models face two fundamental challenges in real-world clinical deployment. First, the acquisition of high-quality expert annotations is highly expensive and time-consuming, which severely limits the scale of labeled data available for supervised learning. Second, lesion phenotypes in medical images often exhibit a continuous spectrum rather than discrete categories, rendering models particularly prone to overfitting under small-sample conditions. 

To address these challenges, we propose a lightweight quantum-enhanced self-supervised learning framework tailored for small-sample medical imaging scenarios. By introducing a two-stage training paradigm, the proposed approach significantly improves the robustness and generalization of feature representations while maintaining high computational efficiency.

Let the original medical image dataset be defined as
\begin{equation}
\mathcal{X} = \{x_i\}_{i=1}^{N}, \quad x_i \in \mathbb{R}^{H \times W},
\end{equation}

where $N$ denotes the number of samples, and $H$ and $W$ represent the spatial resolution of the images. Notably, we do not assume that all samples are associated with reliable annotations. Instead, unlabeled images are treated as a valuable source for mining structural priors inherent to medical imaging data.

In the first stage, the model is trained in a fully self-supervised manner without any manual annotations. A contrastive learning objective is employed to construct a general-purpose feature encoder, defined as
\begin{equation}
f_{\theta}: \mathbb{R}^{H \times W} \rightarrow \mathbb{R}^{d},
\end{equation}

where $\theta$ denotes the network parameters and $d$ is the dimensionality of the embedding space. The primary objective of this stage is not to directly learn classification decision boundaries, but to enforce instance-level contrastive constraints that induce a well-structured geometric distribution in the feature space. As a result, representations of the same anatomical instance remain consistent under different imaging perturbations.

In the second stage, the pretrained encoder obtained from the self-supervised phase is fine-tuned using a small set of labeled samples,
\begin{equation}
\mathcal{D}_{l} = \{(x_i, y_i)\}_{i=1}^{M}, \quad M \ll N,
\end{equation}

through supervised learning. By freezing or partially unfreezing the encoder parameters, the model preserves its general structural perception capability while gradually adapting to specific clinical tasks, such as lesion classification.

The overall network architecture consists of three synergistic components: a lightweight classical feature extractor, a quantum-enhanced feature transformation layer, and a task-specific classification head. This design achieves a systematic balance among parameter efficiency, representational capacity, and training stability.

\subsection{Self-Supervised Contrastive Learning for Medical Imaging}

\subsubsection{Medical Image-Oriented Contrastive Augmentation Mapping}

Unlike natural images, discriminative information in medical images is highly dependent on anatomical continuity, relative organ positioning, and tissue-specific physical properties encoded in grayscale distributions. Overly aggressive augmentation strategies may disrupt critical lesion patterns and consequently mislead the self-supervised learning process.

Motivated by this observation, we construct a medical image-specific stochastic augmentation family in the implementation, defined as
\begin{equation}
\mathcal{T} = \{ t(\cdot; \xi) \mid \xi \sim P(\xi) \},
\end{equation}

where $\xi$ denotes the augmentation parameter set, including rotation angles, cropping ratios, affine perturbations, and intensity variation ranges. For any given input image $x_i$, two distinct views are generated by independently sampling from the same augmentation distribution:
\begin{equation}
\tilde{x}_i^{(1)} = t_1(x_i; \xi_1), \quad
\tilde{x}_i^{(2)} = t_2(x_i; \xi_2).
\end{equation}

This design serves a dual methodological purpose. On the one hand, positive pairs preserve semantic consistency at the anatomical level, ensuring that the learned representations capture stable structural features. On the other hand, controlled perturbations in the pixel space effectively suppress shortcut learning based on low-level textures or imaging noise.

\subsubsection{Geometric Consistency-Driven Contrastive Embedding Learning}

The augmented image views are mapped into a contrastive embedding space through a shared encoder followed by a projection head:
\begin{equation}
z_i^{(k)} =
\frac{g_{\phi}\big(f_{\theta}(\tilde{x}_i^{(k)})\big)}
{\left\| g_{\phi}\big(f_{\theta}(\tilde{x}_i^{(k)})\big) \right\|_2},
\quad k \in \{1,2\},
\end{equation}

where $g_{\phi}(\cdot)$ denotes the projection head network. $\ell_2$ normalization constrains the embeddings onto a unit hypersphere, thereby emphasizing angular similarity as the primary metric.

From this perspective, the self-supervised contrastive objective can be interpreted as maximizing conditional distribution consistency within the embedding space. A temperature-scaled similarity function is introduced as
\begin{equation}
\text{sim}(z_i, z_j) = \frac{z_i^{\top} z_j}{\tau},
\end{equation}

where the temperature parameter $\tau$ controls the concentration of the embedding distribution. Smaller values of $\tau$ emphasize discrimination among hard negative samples, whereas larger values improve training stability.

Finally, by minimizing a conditional cross-entropy loss over positive pairs, the model is encouraged to form a clear and well-separated geometric structure in the feature space, providing a robust foundation for subsequent supervised fine-tuning.
\subsection{Lightweight Quantum-Enhanced Encoder Design}

\subsubsection{Learnable Mapping from Classical Features to Quantum States}

In the classical encoding stage, MobileNetV2 is adopted as the backbone network to substantially reduce the parameter count while preserving sufficient representational capacity. The high-dimensional feature vector extracted by the encoder is denoted as
\begin{equation}
\mathbf{h}_i = f_{\theta}(x_i) \in \mathbb{R}^{d}.
\end{equation}

Given the stringent limitations on the number of qubits under current quantum hardware and simulation conditions, directly embedding high-dimensional classical features into quantum states is computationally infeasible. To address this issue, we introduce a learnable linear compression mapping defined as
\begin{equation}
\mathbf{u}_i = \mathbf{W}_q \mathbf{h}_i + \mathbf{b}_q, \quad \mathbf{u}_i \in \mathbb{R}^{Q},
\end{equation}

where $Q$ denotes the number of qubits. This mapping not only performs dimensionality reduction, but also adaptively selects the most discriminative feature subspace during training, providing a compact yet information-dense representation for subsequent quantum modeling.

The compressed features are then embedded into quantum states via angle encoding:
\begin{equation}
\lvert \psi_i^{(0)} \rangle =
\bigotimes_{q=1}^{Q} R_Y\big(u_{i,q}\big) \lvert 0 \rangle,
\end{equation}

where $R_Y(\cdot)$ denotes a parameterized rotation gate around the $Y$-axis, and $u_{i,q}$ controls the rotation angle of the $q$-th qubit. This operation enables a continuous mapping from classical features to the probability amplitude distribution of quantum states.

\subsubsection{Parameterized Quantum Variational Evolution with Residual Enhancement}

To further capture high-order feature correlations that are difficult to model explicitly using classical neural networks, we construct a multi-layer parameterized quantum variational circuit:
\begin{equation}
\lvert \psi_i^{(L)} \rangle = U(\boldsymbol{\Theta}) \lvert \psi_i^{(0)} \rangle,
\end{equation}

where $U(\boldsymbol{\Theta})$ consists of alternating single-qubit rotation gates and entangling gates, and $\boldsymbol{\Theta}$ represents the set of trainable quantum parameters. Through quantum entanglement, this variational structure introduces non-classical correlations across feature dimensions, enabling expressive modeling of complex feature interactions.

Quantum-enhanced features are obtained by measuring the expectation values of the Pauli-$Z$ operators:
\begin{equation}
\mathbf{q}_i =
\langle \psi_i^{(L)} \rvert \mathbf{Z} \lvert \psi_i^{(L)} \rangle
\in \mathbb{R}^{Q}.
\end{equation}

Finally, a residual fusion strategy is employed to inject the quantum-enhanced features into the classical representation space. A learnable scalar $\alpha$ dynamically controls the contribution of the quantum module:
\begin{equation}
\tilde{\mathbf{h}}_i = \mathbf{h}_i + \alpha \cdot \mathbf{W}_r \mathbf{q}_i.
\end{equation}

This design effectively mitigates training instability caused by the quantum module during early optimization stages and ensures smooth gradient propagation.

\subsection{Supervised Fine-Tuning and Task-Specific Discriminative Modeling}

After completing self-supervised pretraining, a fully connected classification head incorporating Batch Normalization and Dropout is constructed, and end-to-end supervised fine-tuning is performed using the labeled dataset. Through the proposed two-stage training strategy, the model achieves a progressive learning process—from unlabeled structural perception, through quantum-enhanced feature reconstruction, to small-sample discriminative optimization—while maintaining lightweight computational complexity.

This framework is particularly well-suited for clinical medical imaging scenarios characterized by limited annotations and ambiguous class boundaries. The task-specific prediction is given by
\begin{equation}
\hat{y}_i = \mathrm{Softmax}\big(\mathbf{W}_c \tilde{\mathbf{h}}_i + \mathbf{b}_c \big).
\end{equation}

\subsection{Dataset}

The proposed framework was trained and evaluated on publicly available CAG datasets released by the LRSE-Net project, which serves as a benchmark resource for coronary artery stenosis detection. The dataset includes high-quality angiographic images with standardized clinical annotations provided by experienced interventional cardiologists, ensuring both diagnostic reliability and clinical relevance.

Utilizing a public and well-established dataset enables reproducibility and facilitates direct comparison with prior state-of-the-art approaches. All images were processed using a unified preprocessing pipeline. To prevent potential data leakage, dataset splitting was performed at the patient level, with separate subsets allocated for training, validation, and independent testing.

\section{Results}

\subsection{Overall Performance Comparison Across Models}

Figure~1 presents a systematic comparison between the proposed SSL-Quantum model and several classical and lightweight baseline models, including ResNet18, MobileNetV2, EfficientNet-B0, and SimpleCNN, on an independent test set. Multiple evaluation metrics are reported, including Accuracy, AUC, F1 score, Precision, Recall, Sensitivity, and Specificity.

As observed, SSL-Quantum achieves either the best or near-best performance across most core metrics, demonstrating its overall advantage in medical image binary classification tasks. In terms of overall classification accuracy, SSL-Quantum reaches an Accuracy of 0.8083, substantially outperforming ResNet18, MobileNetV2, and EfficientNet-B0. The performance gap is particularly pronounced when compared with the parameter-efficient SimpleCNN. These results indicate that, even without relying on large-scale parameterization or complex architectures, the integration of self-supervised pretraining and quantum-enhanced feature modeling significantly improves the model’s ability to capture key discriminative patterns.

From the perspective of the AUC, which is of particular importance in medical imaging research, SSL-Quantum achieves the highest value of 0.9381, clearly surpassing all baseline models. This suggests a more stable and robust discriminative capability across varying decision thresholds. In contrast, although ResNet18 and EfficientNet-B0 exhibit reasonable baseline performance, they struggle to fully exploit latent discriminative structures under small-sample conditions. These findings highlight the superior global discriminative capacity of the proposed framework, especially in scenarios characterized by class overlap and ambiguous decision boundaries, which are common in clinical imaging data.

\begin{figure}[htbp]
    \centering
    \includegraphics[width=0.75\linewidth]{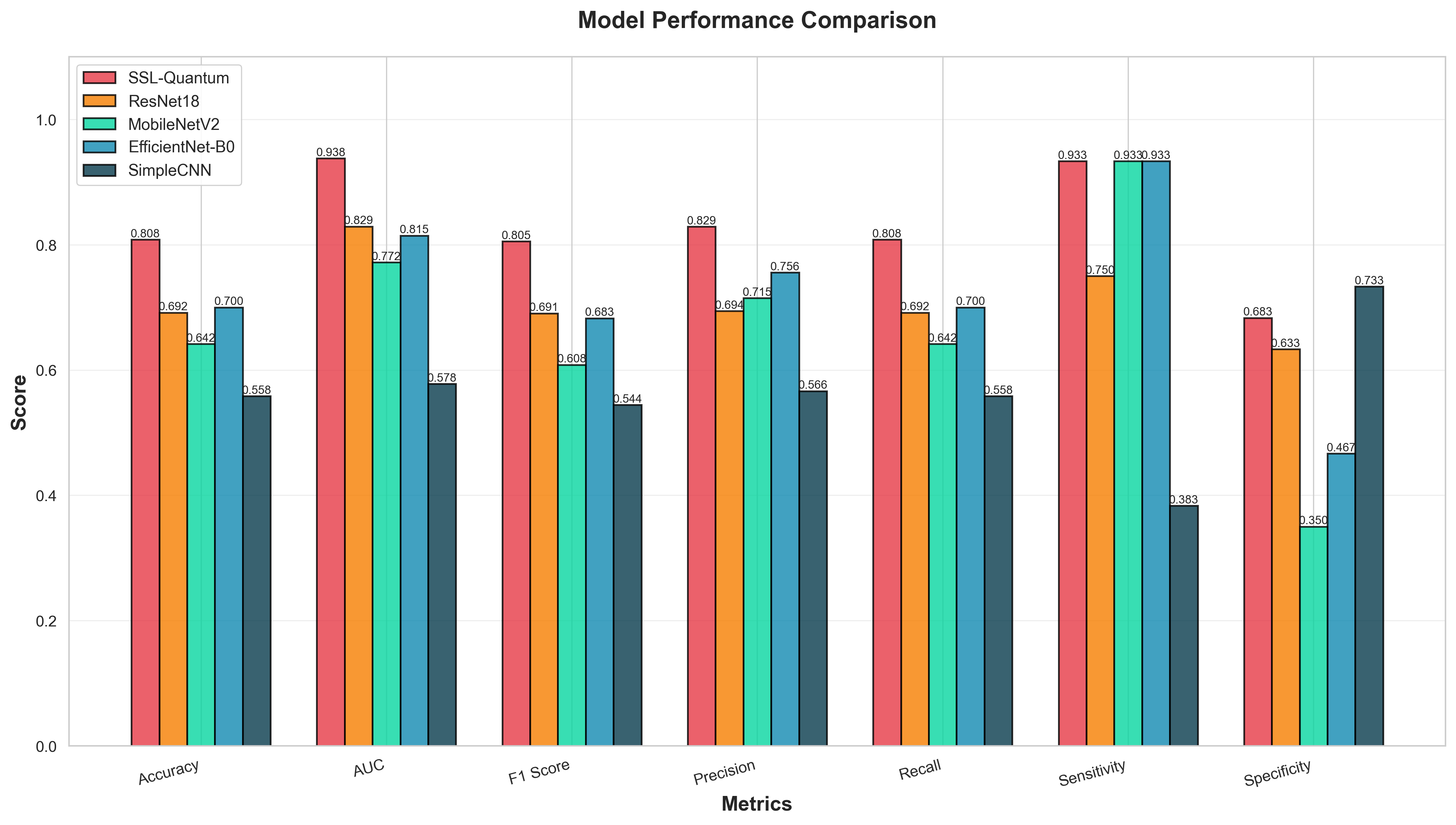}
    \caption{Key Performance Metrics for Models.}
    \label{fig:performance_comparison}
\end{figure}

\subsection{Comprehensive Multi-Metric Performance Visualization}

To further facilitate an intuitive comparison of model behavior across multiple performance dimensions, Figure~2 illustrates radar plots constructed using Accuracy, AUC, F1 score, Precision, Recall, Sensitivity, and Specificity. This visualization emphasizes whether a model exhibits systematic weaknesses rather than excelling in a single isolated metric.

As shown by the overall shape of the radar plots, SSL-Quantum demonstrates the most balanced and outward-expanding polygon, maintaining consistently high performance across nearly all critical dimensions. In particular, the advantages in AUC, F1 score, and Sensitivity—metrics of high relevance to medical imaging tasks—are especially pronounced. This indicates that the proposed model does not achieve performance gains by sacrificing specific aspects, but instead realizes a coordinated optimization of discriminative power, stability, and robustness.

By comparison, MobileNetV2 and EfficientNet-B0 show relatively strong performance in Sensitivity but exhibit noticeable contraction in Specificity and overall Accuracy, suggesting an imbalance between positive and negative sample discrimination. The radar plot of SimpleCNN displays a markedly contracted shape, further confirming its limited representational capacity for complex medical imaging tasks.

\begin{figure}[htbp]
    \centering
    \includegraphics[width=0.5\linewidth]{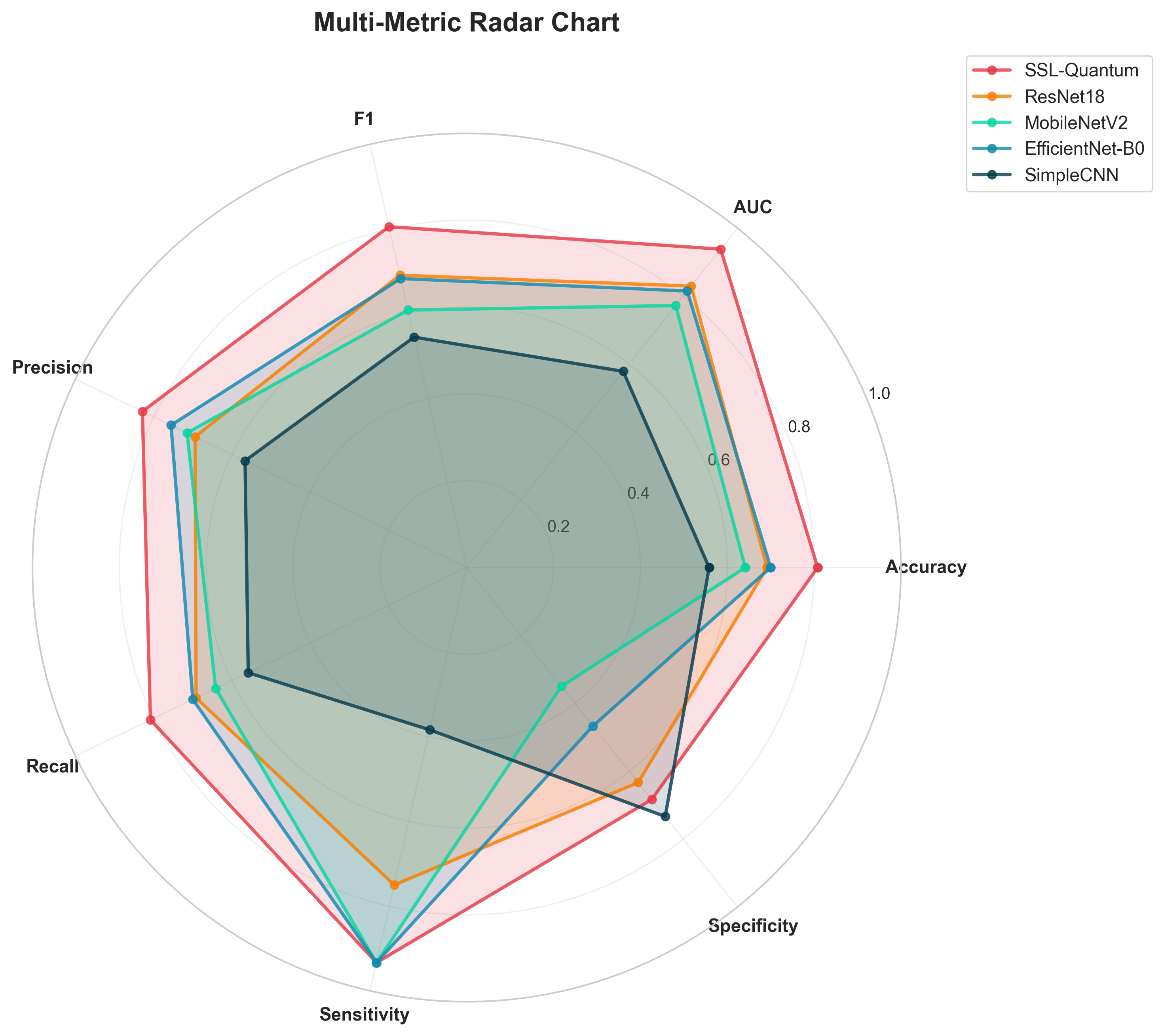}
    \caption{Multi-Dimensional Performance Metrics.}
    \label{fig:performance_comparison}
\end{figure}

\subsection{Analysis of Classification Behavior via Confusion Matrices}

Figure~3 presents the confusion matrices of all evaluated models on the test set, providing deeper insights into error distribution patterns between positive and negative classes. In medical imaging applications, accuracy alone often fails to fully characterize model behavior, whereas confusion matrices explicitly reveal the clinical implications of false positives and false negatives.

SSL-Quantum exhibits a relatively balanced error distribution, achieving high numbers of both true positives and true negatives, while maintaining a notably lower false negative rate compared to ResNet18 and SimpleCNN. This characteristic is particularly critical in clinical contexts, as false negatives correspond to missed detections of potential lesions and are associated with higher clinical risk.

In contrast, MobileNetV2 and EfficientNet-B0 achieve relatively high true positive counts but suffer from elevated false positive rates, indicating a tendency toward over-sensitivity. While this may improve detection rates, it comes at the cost of reduced specificity and may lead to unnecessary follow-up examinations or invasive procedures in real-world practice. Overall, SSL-Quantum achieves a more favorable balance between sensitivity and specificity, aligning more closely with practical clinical decision-making requirements.

\begin{figure}[htbp]
    \centering
    \includegraphics[width=0.75\linewidth]{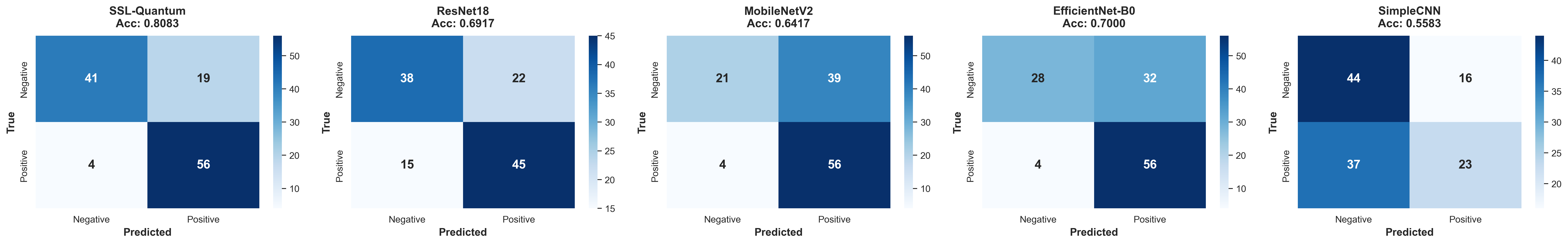}
    \caption{Comparison Across Baseline and Proposed Models.}
    \label{fig:performance_comparison}
\end{figure}

\subsection{Decision Threshold Stability Analysis}

Figure~4 illustrates the ROC curves and corresponding AUC values of different models on the test set, enabling an assessment of discriminative stability across varying decision thresholds. From the curve profiles, SSL-Quantum consistently dominates the other models across nearly the entire false positive rate range and maintains a high true positive rate even in low false positive regions.

This property is of substantial clinical importance, as real-world diagnostic systems often operate under stringent false positive constraints to avoid overdiagnosis. The ability of SSL-Quantum to preserve high sensitivity under such conditions indicates that the learned representations exhibit clearer separation in the decision space, rather than relying on a narrow range of threshold-dependent optimality. In contrast, the ROC curve of SimpleCNN approaches that of a random classifier, with an AUC of only 0.5781, underscoring its limited capacity for modeling complex medical image features. Although ResNet18 and EfficientNet-B0 demonstrate moderate discriminative ability, they remain inferior to SSL-Quantum in both overall curve coverage and stability.

\begin{figure}[htbp]
    \centering
    \includegraphics[width=0.5\linewidth]{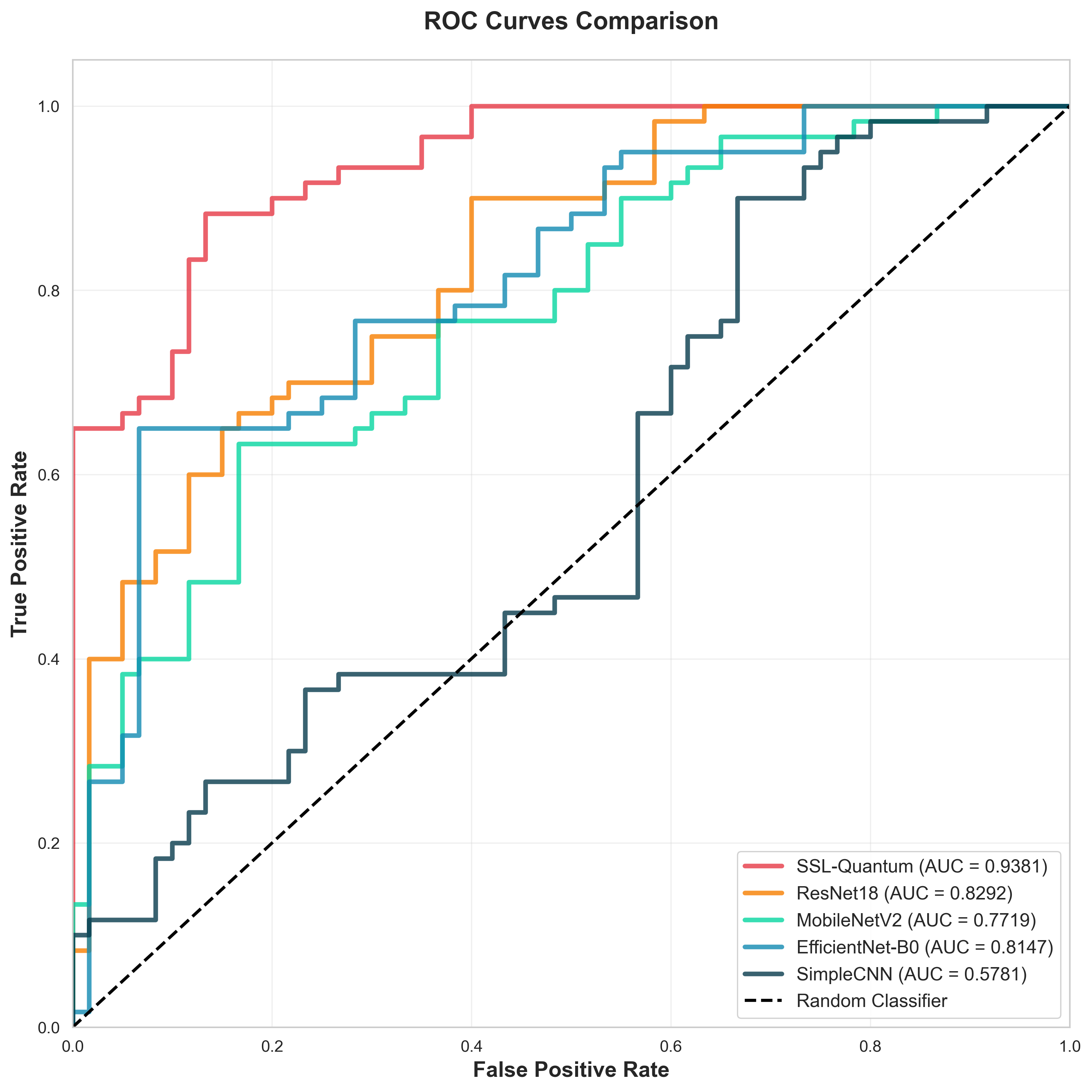}
    \caption{ROC Curves and AUC for Comparative Model Assessment.}
    \label{fig:performance_comparison}
\end{figure}

\subsection{Quantitative Performance Summary}

Table~\ref{tab:performance_comparison} summarizes all evaluation metrics for the tested models on the independent test set, providing quantitative support for the preceding visual analyses. As shown, SSL-Quantum consistently outperforms baseline models on comprehensive metrics such as Accuracy, AUC, and F1 score, reflecting a more stable and reliable overall classification performance.

It is noteworthy that while SSL-Quantum achieves sensitivity comparable to certain lightweight models, its specificity is markedly higher than that of MobileNetV2 and EfficientNet-B0. This indicates that the proposed approach effectively controls false positive errors while maintaining a high detection rate. Such a performance profile is particularly desirable in medical image-assisted diagnosis, as it directly affects the clinical acceptability and downstream decision-making impact of AI systems.

\begin{table}[t]
\centering
\caption{Quantitative comparison of model performance on the test set.}
\label{tab:performance_comparison}
\begin{tabular}{lccccccc}
\hline
Model & Accuracy & AUC & F1 & Precision & Recall & Sensitivity & Specificity \\
\hline
SSL-Quantum & 0.8083 & 0.9381 & 0.8053 & 0.8289 & 0.8083 & 0.9333 & 0.6833 \\
ResNet18 & 0.6917 & 0.8292 & 0.6906 & 0.6943 & 0.6917 & 0.7500 & 0.6333 \\
MobileNetV2 & 0.6417 & 0.7719 & 0.6083 & 0.7147 & 0.6417 & 0.9333 & 0.3500 \\
EfficientNet-B0 & 0.7000 & 0.8147 & 0.6827 & 0.7557 & 0.7000 & 0.9333 & 0.4667 \\
SimpleCNN & 0.5583 & 0.5781 & 0.5444 & 0.5665 & 0.5583 & 0.3833 & 0.7333 \\
\hline
\end{tabular}
\end{table}

Overall, these results demonstrate that the proposed SSL-Quantum framework consistently outperforms conventional supervised models and lightweight CNN architectures across multiple key metrics, even under GPU-free and parameter-constrained experimental settings. This validates the effectiveness and practical potential of combining self-supervised contrastive learning with quantum-enhanced feature modeling for small-sample medical imaging applications.

\section{Discussion}

From the perspective of deep integration between medical imaging and artificial intelligence, this study systematically evaluated the effectiveness and feasibility of a lightweight hybrid learning framework that combines self-supervised contrastive learning with a quantum-enhanced module for medical image binary classification. Through comprehensive comparisons with multiple classical and mainstream lightweight convolutional neural network architectures, the proposed approach consistently demonstrated stable and superior performance across a range of key evaluation metrics. These findings not only validate the technical soundness of the proposed method, but also provide a substantive response to several fundamental challenges in contemporary medical imaging AI research, including the scarcity of high-quality annotated data, limitations in computational resources, and insufficient generalization under complex clinical scenarios.

From the perspective of learning paradigms, the experimental results clearly indicate that self-supervised contrastive learning in medical image analysis should not be regarded merely as a form of data augmentation, but rather as a principled mechanism for fundamentally improving feature representation quality. Compared with traditional training strategies that rely exclusively on supervised signals, the self-supervised pretraining stage constructs positive and negative sample pairs to guide the model toward learning more discriminative and stable latent feature distributions from unlabeled data. In this study, this advantage manifests in the observation that, even under identical network capacity and similar training configurations, models equipped with self-supervised pretraining significantly outperform their purely supervised counterparts in downstream classification tasks.

This finding is particularly relevant from a medical imaging perspective. Medical images are typically characterized by high structural homogeneity, complex noise patterns, and subtle phenotypic differences between pathological and normal tissues. When training relies solely on limited annotated samples, models are prone to overfitting superficial textures or locally salient contrast regions. In contrast, self-supervised contrastive learning enforces representation consistency across multiple views and perturbations of the same image, encouraging the model to focus on structural and semantic information rather than incidental low-level visual cues. As a result, the learned representations exhibit enhanced robustness and maintain discriminative power even under small-sample conditions or mild distribution shifts in test data.

At an empirical level, the results of this study support a perspective that has attracted increasing attention yet remains subject to debate: quantum-inspired computational paradigms can provide tangible benefits to deep learning models even under current classical simulation conditions. In particular, the parameterized quantum circuits introduced in this work are not employed as standalone classifiers, but are instead designed as embedded feature enhancement modules that are tightly coupled with classical neural networks in a hybrid architecture. This design choice effectively circumvents the limitations of contemporary quantum hardware, such as high noise levels and restricted scalability, while leveraging the non-linear mapping capacity of quantum states in high-dimensional Hilbert spaces.

Importantly, the incorporation of the quantum-enhanced module does not introduce training instability or performance fluctuations. On the contrary, consistent performance improvements are observed across multiple metrics, especially those of high clinical relevance such as AUC, F1 score, and sensitivity. These observations suggest that the quantum component does not merely increase model complexity, but instead provides a meaningful complementary representation that enriches the classical feature space and enables the model to better capture complex correlation structures inherent in medical imaging data.

Beyond predictive performance, computational efficiency and deployment feasibility are critical considerations in real-world medical imaging applications. A key finding of this study is that competitive classification performance can be achieved without reliance on large-scale parameterization or high-end computational hardware, provided that appropriate learning paradigms and feature enhancement strategies are employed. Compared with lightweight yet widely adopted architectures such as ResNet18 and EfficientNet-B0, the proposed framework achieves a more balanced performance profile while maintaining relatively low computational complexity. This has direct implications for resource-constrained clinical environments, primary healthcare institutions, and mobile or edge-based medical imaging systems.

Conventional wisdom often associates high-performance models with deep architectures and large parameter budgets. In contrast, the results of this study demonstrate that the integration of self-supervised learning and quantum-enhanced feature modeling enables a more favorable trade-off among model size, training cost, and predictive performance. This insight provides a promising alternative pathway for the practical deployment and engineering translation of medical imaging AI systems.

It is worth emphasizing that this work does not focus on improving a single evaluation metric in isolation, but instead systematically analyzes model behavior across multiple complementary performance indicators. From Accuracy and AUC to F1 score, sensitivity, and specificity, the proposed method exhibits consistent advantages across diverse dimensions. Such multi-metric consistency is particularly important in medical imaging applications, where clinical decision-making depends on the balance between different types of errors rather than the maximization of a single scalar metric. Notably, the proposed framework maintains high sensitivity while avoiding substantial degradation in specificity, thereby reducing the risk of missed diagnoses without inducing excessive false alarms. This balanced performance profile enhances clinical trust and reflects greater stability and robustness in decision boundary learning.

Despite these encouraging results, several limitations should be acknowledged. First, the quantum-enhanced module is implemented using classical simulation, and its scale and complexity are constrained by current computational resources. As such, the full potential advantages of quantum hardware cannot yet be fully realized. Second, this study focuses primarily on binary classification tasks, and the effectiveness of the proposed framework in multi-class or more complex medical imaging scenarios remains to be systematically investigated. Third, the dataset used in this work is derived from a relatively limited source, and further validation across multi-center, multi-vendor datasets is necessary to rigorously assess generalization performance.

Future research directions are therefore multifaceted. On the one hand, continued advances in quantum hardware and hybrid computing frameworks may enable deeper exploration of quantum–classical synergistic mechanisms. On the other hand, self-supervised learning strategies themselves offer substantial design flexibility; incorporating modality-specific priors and anatomically informed contrastive objectives may further enhance performance and interpretability across diverse medical imaging modalities.

\section{Conclusion}

This study proposes a lightweight hybrid learning framework for medical image classification that integrates self-supervised contrastive learning with quantum-enhanced feature modeling, and systematically validates its effectiveness and feasibility under resource-constrained computational settings. The experimental results demonstrate that embedding self-supervised pretraining and parameterized quantum circuits into a lightweight deep network can substantially improve discriminative performance and predictive stability without imposing significant additional computational overhead.

From a methodological perspective, this work introduces a novel paradigm for medical image representation learning, in which self-supervised learning serves as the foundation for robust structural feature extraction, while quantum-enhanced modeling provides complementary expressiveness for capturing complex feature correlations. From a practical standpoint, the results offer empirical evidence that quantum computing concepts—when implemented in a hybrid and simulation-compatible manner—can already function as a meaningful augmentation to classical deep learning models at the current stage of technological development.

More importantly, this study underscores that, in real-world clinical deployment scenarios, model design must jointly consider predictive performance, resource constraints, and deployment feasibility. By achieving a favorable balance among these factors, the proposed framework facilitates a transition of medical imaging AI systems from proof-of-concept demonstrations toward clinically viable solutions.

Overall, this work provides a reproducible methodological framework and compelling empirical evidence for the synergistic integration of self-supervised learning and quantum machine learning in medical imaging. It lays a solid foundation for the future development of efficient, scalable, and innovative medical artificial intelligence models that are better aligned with practical clinical needs.

\bibliographystyle{unsrt}
\bibliography{references}

@article{1,
  title={Self-supervised learning for medical image analysis using image context restoration},
  author={Chen, Liang and Bentley, Paul and Mori, Kensaku and Misawa, Kazunari and Fujiwara, Michitaka and Rueckert, Daniel},
  journal={Medical image analysis},
  volume={58},
  pages={101539},
  year={2019},
  publisher={Elsevier}
}

@article{2,
  title={Self-supervised learning in medicine and healthcare},
  author={Krishnan, Rayan and Rajpurkar, Pranav and Topol, Eric J},
  journal={Nature Biomedical Engineering},
  volume={6},
  number={12},
  pages={1346--1352},
  year={2022},
  publisher={Nature Publishing Group UK London}
}

@inproceedings{3,
  title={Building a General SimCLR Self-Supervised Foundation Model Across Neurological Diseases to Advance 3D Brain MRI Diagnoses},
  author={Kaczmarek, Emily and Szeto, Justin and Nichyporuk, Brennan and Arbel, Tal},
  booktitle={Proceedings of the IEEE/CVF International Conference on Computer Vision},
  pages={1310--1319},
  year={2025}
}

@article{4,
  title={RS-MOCO: A deep learning-based topology-preserving image registration method for cardiac T1 mapping},
  author={Huang, Chiyi and Sun, Longwei and Liang, Dong and Wang, Haifeng and Zeng, Hongwu and Zhu, Yanjie},
  journal={Computers in Biology and Medicine},
  volume={184},
  pages={109442},
  year={2025},
  publisher={Elsevier}
}

@article{5,
  title={Self-Supervised Learning with BYOL for Non-Alcoholic Fatty Liver Disease Diagnosis Using Ultrasound Imaging},
  author={Buktash, Ali and G{\"o}r{\"u}r, Abd{\"u}l Kadir},
  journal={Signal, Image and Video Processing},
  volume={19},
  number={15},
  pages={1--10},
  year={2025},
  publisher={Springer}
}

@article{6,
  title={A hybrid quantum-classical approach for liver disease detection using quantum machine learning},
  author={Donaire, Laura Mar{\'\i}a and Ortega, Gloria and Orts, Francisco and Garz{\'o}n, Ester Mart{\'\i}n and Filatovas, Ernestas},
  journal={Engineering Applications of Artificial Intelligence},
  volume={164},
  pages={113240},
  year={2026},
  publisher={Elsevier}
}

@article{7,
  title={Optimizing biomedical waste generation modeling using quantum machine learning and economic development indicators},
  author={Aliyu, Usman U and Mahmoud, Ismail A and Mati, Sagir and Chaki, Sukalpaa and Sulaiman, Tukur Abdulkadir and Usman, AG and Abba, Sani I},
  journal={Biomass and Bioenergy},
  volume={204},
  pages={108312},
  year={2026},
  publisher={Elsevier}
}

@incollection{8,
  title={Drug Discovery with Quantum Machine Learning},
  author={Pyrkov, Alexey and Aliper, Alex and Bezrukov, Dmitry and Zhavoronkov, Alex},
  booktitle={Applied Artificial Intelligence for Drug Discovery: From Data-Driven Insights to Therapeutic Innovation},
  pages={289--343},
  year={2026},
  publisher={Springer}
}

@article{9,
  title={A hybrid quantum-neural network for heart disease classification},
  author={Verdone, Alessio and Succetti, Federico and Ceschini, Andrea and Rosato, Antonello and Fioravanti, Alessio and Panella, Massimo},
  journal={Biomedical Signal Processing and Control},
  volume={113},
  pages={109185},
  year={2026},
  publisher={Elsevier}
}

@article{10,
  title={Applications of Quantum-Inspired Soft Computing for Intelligent Data Processing in Real-Life Scenarios},
  author={Suyal, Priyanka and Gola, Kamal Kumar and Chakraborty, Camellia and Kanauzia, Rohit and Suyal, Mohit and Mridula},
  journal={Quantum-Inspired Approaches for Intelligent Data Processing},
  pages={223--257},
  year={2026},
  publisher={Wiley Online Library}
}

@article{11,
  title={Quantum-inspired adaptive feature fusion for highly accurate brain tumor classification in MRI using deep learning},
  author={Meenal, T and Asokan, R},
  journal={Biomedical Signal Processing and Control},
  volume={112},
  pages={108694},
  year={2026},
  publisher={Elsevier}
}

@article{12,
  title={Investigating Quantum Feature Maps in Quantum Support Vector Machines for Lung Cancer Classification},
  author={Toufah, A and Kadim, MA and El Hafidi, Moulay Youssef},
  journal={Journal of Artificial Intelligence Research and Innovation},
  volume={1},
  number={1},
  pages={100--110},
  year={2026}
}

@article{13,
  title={ARTIFICIAL INTELLIGENCE IN CLINICAL MEDICINE: CURRENT APPLICATIONS, CHALLENGES, AND FUTURE DIRECTIONS},
  author={Alieva, Zarnigor and Egamberdieva, Gulchekhra},
  journal={Journal of Clinical and Biomedical Research},
  volume={1},
  number={1},
  pages={46--50},
  year={2026}
}

@article{14,
  title={Knowledge fusion in deep learning-based medical vision-language models: A review},
  author={Xu, Dexuan and Chen, Yanyuan and Chai, Zhongyan and Xiao, Yifan and Yan, Yandong and Ding, Weiping and Wang, Hanpin and Jin, Zhi and Jiao, Wenpin and Yue, Weihua and others},
  journal={Information Fusion},
  volume={125},
  pages={103455},
  year={2026},
  publisher={Elsevier}
}

@article{15,
  title={Deep learning-based bone marrow cytology classification: A solution to class imbalance},
  author={Bonab, Zahra Asgharzadeh and Shamekhi, Sina and Talebi, Mehdi},
  journal={Biomedical Signal Processing and Control},
  volume={111},
  pages={108247},
  year={2026},
  publisher={Elsevier}
}

@article{16,
  title={AdveDiffNet: adversarial diffusion network for unbalanced melanoma diagnosis},
  author={Fu, Yu and Liu, Chao and Wang, Shaoqiang and Xia, Hui},
  journal={Biomedical Signal Processing and Control},
  volume={113},
  pages={108952},
  year={2026},
  publisher={Elsevier}
}

\end{document}